%% file: main.tex
\documentclass{article}
\usepackage{microtype}
\usepackage{graphicx}
\usepackage{booktabs}
\usepackage{amsmath,amssymb}
\usepackage{xcolor}
\usepackage{hyperref}
\usepackage[acronym]{glossaries}

\usepackage[accepted]{mlsys2025_arxiv} 

\graphicspath{{figs/}}
\newcommand{\degC}{^{\circ}\mathrm{C}}
\newcommand{\E}{\mathbb{E}}
\newcommand{\R}{\mathbb{R}}
\DeclareMathOperator{\clip}{clip}
\definecolor{teal}{HTML}{2A9D8F}
\newcommand{\ev}[1]{\textcolor{teal}{\textbf{[#1]}}}

\mlsystitlerunning{From Data to Valves: A Self-Learning, Production-Grade Edge Stack for Sustainable HVAC Control}

\begin{document}

\include{glossary}

\twocolumn[
\mlsystitle{From Data to Valves: A Self-Learning, Production-Grade Edge Stack for Sustainable HVAC Control}

\mlsyssetsymbol{equal}{*}
\begin{mlsysauthorlist}
\mlsysauthor{Abbas Raza Ali}{equal,abbas}
\mlsysauthor{Muhammad Hamza Rehman}{hamza}
\mlsysauthor{Xu Li}{alice}
\mlsysauthor{Muhammad Ajmal Siddiqui}{ajmal}
\end{mlsysauthorlist}

\mlsysaffiliation{abbas}{Citigrounp Inc., London, United Kingdom}
\mlsysaffiliation{hamza}{National University of Computing \& Emerging Sciences, Islamabad, Pakistan}
\mlsysaffiliation{alice}{Alibaba Cloud, Amsterdam, Netherlands}
\mlsysaffiliation{ajmal}{Ernst \& Young LLP, London, United Kingdom}

\mlsyscorrespondingauthor{Abbas Raza Ali}{abbas.raza.ali@gmail.com}

\mlsyskeywords{Edge, online learning, HVAC control, Timeseries Forecasting, Reinforement Learning, MLOps}

\vskip 0.3in

\begin{abstract}
Learned HVAC controllers rarely leave the simulator, and those that do are typically trained offline, frozen at deployment and retrained from the cloud. We present a production stack in which three learning components run on one NVIDIA Jetson Orin gateway per building, adapt continuously from the building's own data, and verify their own effect. (a) A compact transformer-encoder forecaster, one per equipment group with learned per-unit embeddings, predicts every air-side unit's next-cycle temperature and learns prequentially, one gradient step per five-minute cycle, behind five independent gates that bound the influence of any corrupt reading. (b) A distributional soft actor-critic agent, one shared policy across unit types, converts the measured state into a chilled-water valve position under a structural safety envelope that no learned component can soften; its second generation learns nightly on the device from recorded transitions alone, with a behaviour anchor and critic calibration terms, is scored by four model-free metrics and promoted by atomic hot-swap. (c) A change-point regression baseline held to ASHRAE Guideline 14 supplies the counterfactual against which savings are certified. On eighteen months of data from a hotel the forecaster converges from random weights within a week and reaches 0.37, 0.39 and 0.11\,$\degC$ MAE for air-handling, fresh-air and fan-coil units with no retraining event; on a different hotel the baseline reproduces a 46-day held-out period to within 0.07\,\%. The RL agent has completed its in-simulator campaign and runs in shadow on 104 units; a sixty-day replay campaign establishes fitness to take control and shows that outperformance can only be established in the active pilot; we report measured latency and memory of every service on a shared 15\,GB device, the systems lessons of the campaign, and precisely what the evidence does and does not yet establish.
\end{abstract}
]

\printAffiliationsAndNotice{}

\section{Introduction}
Buildings account for roughly 30\,\% of global final energy and HVAC is their largest end use \cite{iea2023buildings,perezlombard2008review}. In hot climates the chilled-water valve of each air-handling unit (AHU), fresh-air handling unit (FAHU) and fan-coil unit (FCU) is the actuator through which cooling is committed, and building management systems (BMS) drive it with a proportional-integral loop tuned once at commissioning. Deep reinforcement learning (RL) for HVAC has a decade of simulation results \cite{wei2017drl,wang2020rlbuildings,yu2021review}, but a recent community statement lists the simulation-to-deployment gap, per-building modelling cost and the absence of standard evaluation among its ten open questions \cite{nagy2023ten}. The few real deployments are policies trained offline and updated from the cloud \cite{zhang2018practical,luo2022deepmind}, and the collect-train-deploy-retrain pattern carries well-documented operational debt \cite{sculley2015debt}.

\begin{figure*}[t]
\centering\includegraphics[width=0.92\textwidth]{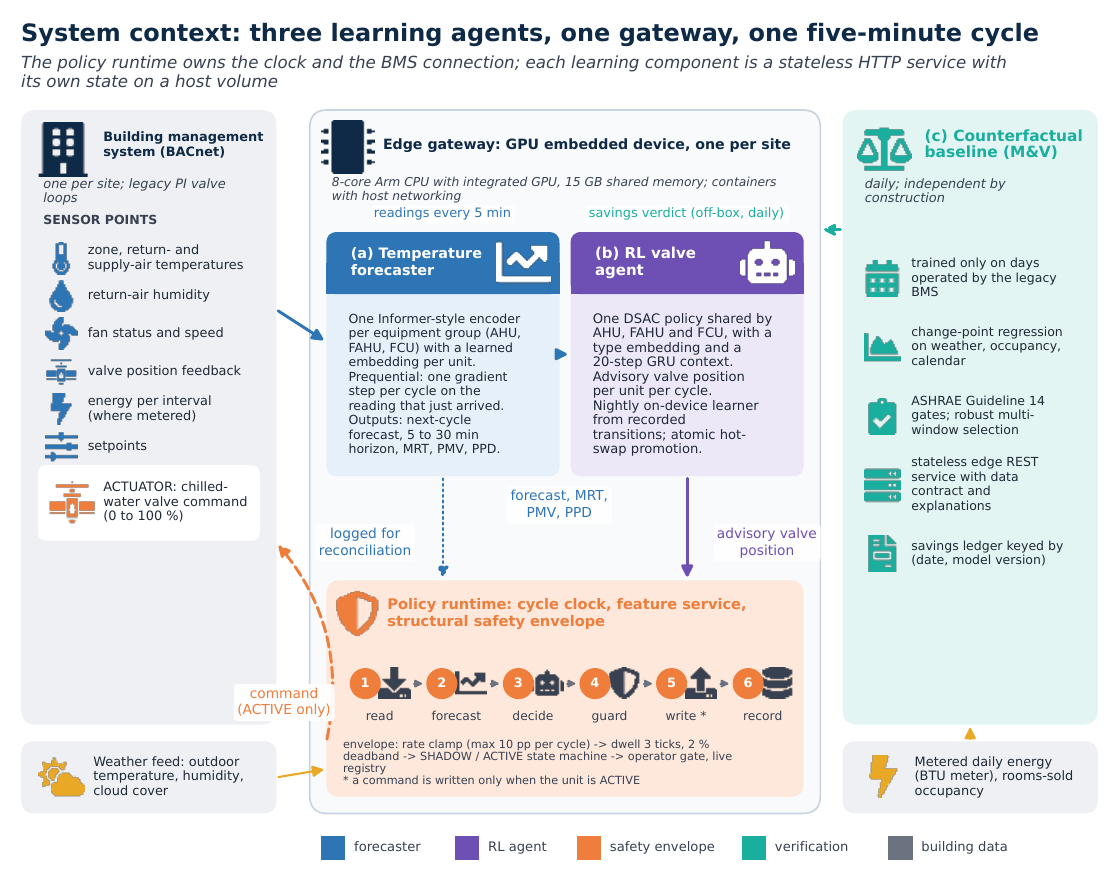}
\caption{System context: three learning agents on one gateway, one five-minute cycle. The policy runtime owns the clock and the BMS connection; the three learning components are stateless HTTP services. BMS writes exist only in ACTIVE mode; the savings verdict closes the loop off-box.}
\label{fig:stack}
\end{figure*}

This paper documents, in the spirit of the industrial track, a system that was built, deployed and measured. Its design principle is one sentence: \emph{learn from the building's own data, on the device, continuously, and make that learning safe and verifiable}. Three cooperating learning components (Figure~\ref{fig:stack}) answer three questions on one five-minute cycle: where is each space heading (a forecaster), what should each valve do (an RL agent) and did it help (a counterfactual baseline). The forecaster has no offline training phase; the RL agent, in its second generation, has no simulator; the baseline uses no signal from the other two.

\textbf{Contributions.} (1) A prequential, fleet-shared forecaster with a fixed-size three-day context memory and a provable bound on the influence of any single corrupt reading, validated on eighteen months of five-minute data (Sections~\ref{sec:fc}, \ref{sec:res}). (2) A simulator-free, on-device RL controller with a structural safety envelope, a nightly off-policy learner using behaviour anchoring and critic calibration, and atomic hot-swap promotion (Section~\ref{sec:rl}). (3) An auditable Guideline-14 baseline served at the edge with a savings ledger (Section~\ref{sec:mv}). (4) A four-timescale learning hierarchy, a coupling ladder between forecaster and controller, and an evidence ladder stating what each kind of evaluation can establish (Sections~\ref{sec:sys}, \ref{sec:eval}). (5) Measured deployment characteristics and transferable systems lessons (Sections~\ref{sec:res}, \ref{sec:lessons}). We tag claims as \ev{Established} (prior work), \ev{Measured} (real site data through the production code path), \ev{Simulated} (inside a data-driven surrogate) or \ev{Proposed}.

\textbf{What we do not claim.} Measured comfort or energy outcomes of the RL agent on a real building are not yet established: both generations have run only in shadow, in which advisories are logged but never executed. The staged path from ``fit to take control'' to certified savings is laid out in Section~\ref{sec:limits}.

\section{Background}
\textbf{Learned building control.} MPC \cite{afram2014mpc,drgona2020mpc} and RL \cite{yu2021review,ali2019metarl} are the alternatives to rule-based control; simulation benchmarks \cite{blum2021boptest,jimenezraboso2021sinergym,vazquezcanteli2019citylearn} make results comparable but also document why they transfer poorly \cite{nagy2023ten}. Deployments and near-deployment studies include a calibrated-simulator policy on a radiant system \cite{zhang2018practical}, a differentiable-MPC policy warm-started from logs \cite{chen2019gnurl}, RL chiller control behind hard constraints \cite{luo2022deepmind}, imitation plus online RL in simulation \cite{solinas2024online} and offline RL from BMS logs \cite{blad2022offline}. Table~\ref{tab:related} positions this work. \textbf{Online learning.} Prequential evaluation \cite{dawid1984prequential,gama2013evaluating}, online learning \cite{hoi2021online} and concept drift \cite{gama2014drift,lu2019drift} supply the forecaster's protocol; compact models often match large transformers on time series \cite{ali2018crossdomain}. \textbf{Offline and safe RL.} SAC \cite{haarnoja2018sac}, IQN critics \cite{dabney2018iqn,duan2022dsac}, behaviour regularisation \cite{fujimoto2021td3bc}, conservative and calibrated critics \cite{nakamoto2023calql}, fitted Q evaluation \cite{le2019fqe,voloshin2021ope} and shielding \cite{alshiekh2018shielding,garcia2015safe} are the ingredients of the controller. \textbf{M\&V.} IPMVP Option C and ASHRAE Guideline 14 \cite{ipmvp2022,ashrae14}, change-point models \cite{kissock2003imt} and the accuracy of automated M\&V \cite{granderson2016accuracy} ground the baseline. \textbf{Agentic AI.} Language-model agents are characterised by perception, memory, planning, tool use, reflection and governance \cite{shavit2023governing}; a systematic review of 66 studies finds language models research-stage for autonomous HVAC control and supports them as semantic and workflow layers around physics- and data-driven controllers \cite{neubauer2026llmhvac}. The stack presented here is the physical loop such a layer would supervise.

\begin{table}[t]
\caption{Representative learned HVAC controllers and this work.}
\label{tab:related}
\vskip 0.1in
\centering\footnotesize
\resizebox{\columnwidth}{!}{\begin{tabular}{@{}lcccc@{}}
\toprule
System & Simulator & Deployed & Online & M\&V \\
\midrule
Zhang \& Lam \cite{zhang2018practical} & yes & frozen & no & no \\
Gnu-RL \cite{chen2019gnurl} & no & lab & yes & no \\
DeepMind \cite{luo2022deepmind} & learned & frozen & periodic & no \\
Solinas et al.\ \cite{solinas2024online} & yes & no & yes & no \\
Blad et al.\ \cite{blad2022offline} & no & no & no & no \\
\textbf{This work (v2)} & \textbf{none} & shadow, 104 & \textbf{nightly} & \textbf{G14} \\
\bottomrule
\end{tabular}}
\vskip -0.1in
\end{table}

\section{System design}\label{sec:sys}
\textbf{Sites and hardware.} Two hotels: hotel-1 (RR; 9 AHUs, 4 FAHUs, about 200 FCUs, exports from September 2024) and hotel-2 (HYI; 545 rooms, 125 units, a daily main BTU meter and an independently reviewed M\&V plan). One Advantech MIC-713 gateway (Jetson Orin NX, 8-core Cortex-A78AE, 15\,GB, JetPack 6.1, CUDA 12.6) per site runs every component as a container with host networking, shared with other tenants.

\textbf{Policy runtime and contracts.} A separate policy-runtime service owns the five-minute clock and is the only process that talks to the BMS. Each cycle it posts a group's readings and weather to the forecasting service, posts the same envelope plus the forecaster's response to the RL service, receives an advisory that has already passed the RL service's safety envelope (rate clamp, dwell guard, shadow state machine), and writes to the BMS only for units whose registry status is \emph{live}. Once per completed day it calls the baseline and records the counterfactual beside the metered energy. Every learning component is a timer-less HTTP endpoint driven by the policy runtime's clock, with per-unit state persisted atomically on the device; offline replay of a historical export constructs the same one-row-per-unit table a live request is flattened into and calls the same step function, which is what makes every offline number below a production number.

\textbf{Design principles.} P1 learn on the device from the building's own data; P2 one model per group, specialised per unit through learned entity embeddings \cite{guo2016entity}; P3 predict, then learn, with one cycle of delay, on one code path; P4 safety is structural, not learned; P5 bounded influence of any single input; P6 verify with an instrument the system cannot touch.

\textbf{Four timescales.} Learning and verification run on four clocks that never block on each other: the control cycle (forecaster step, advisory, recorded transition, safety envelope), the night (RL learner, model-free scoring, promotion), the day (counterfactual kWh, savings ledger) and the quarter (baseline retrain on BMS-operated days). This is what lets each component fail, pause or roll back without stopping the others, and why the device's compute is partitioned by time rather than by contention.

\section{Online-adaptive forecasting}\label{sec:fc}
\textbf{Prequential objective.} For unit $u$ with features $x_u(t_n)\in\R^F$, embedding $e_u\in\R^E$ and target $y_u$ (return-air temperature for AHUs, zone temperature for FCUs, supply-air temperature for FAHUs), the shared model issues $\hat y_u(t_{n+1}) = f_\theta(x_u(t_n), e_u)$; when cycle $n{+}1$ arrives, the loss $\frac1B\sum_u[\hat y_u(t_{n+1})-y_u(t_{n+1})]^2$ over units present in both cycles (matched by identifier) drives exactly one optimiser step on $\theta$ and the embeddings before the next forecast. The forecast is carried on the autograd graph across the inter-cycle gap. No train/test split, replay buffer, epoch or scheduler exists; the per-cycle error sequence is both training signal and evaluation record \ev{Proposed}.

\textbf{Context memory.} Each forward pass consumes a single vector, but the model conditions on three days of each unit's history through a time-of-day slot memory $m_{u,s,k}\leftarrow\alpha x+(1-\alpha)m_{u,s,k}$ with $\alpha=2/(N_{\text{days}}+1)$, $N_{\text{days}}=3$, per unit, sensor and five-minute slot (288 per day), which imputes missing sensors from the unit's own diurnal pattern and is never updated by imputed values, and through the continuously updated weights themselves. A missing target is replaced by the unit's last forecast, never zero. Temperatures are clipped to per-group physical ranges, other columns to $\mu\pm z\sigma$ from Welford statistics on real readings only, and all columns are scaled by fixed configured bounds so that offline and live share constants with no leakage.

\textbf{Model.} A compact Informer-style encoder \cite{vaswani2017attention,sokol2026assortment,ali2014reutilization}: the $F{+}E$ input is projected to $d_{\text{model}}$, passed through post-norm encoder blocks whose attention acts across the $B$ units of the group in the batch (sequence length one, no positional encoding), and reduced to one scalar per unit. New units are registered on first sight with a mean-initialised embedding row added to the optimiser; onboarding is a configuration change. The deployed size is $d_{\text{model}}=32$, 2 heads, 1 layer, $d_{\text{ff}}=128$: about 14\,k parameters per group plus an 8-dim row per unit.

\begin{figure}[t]
\centering\includegraphics[width=\columnwidth]{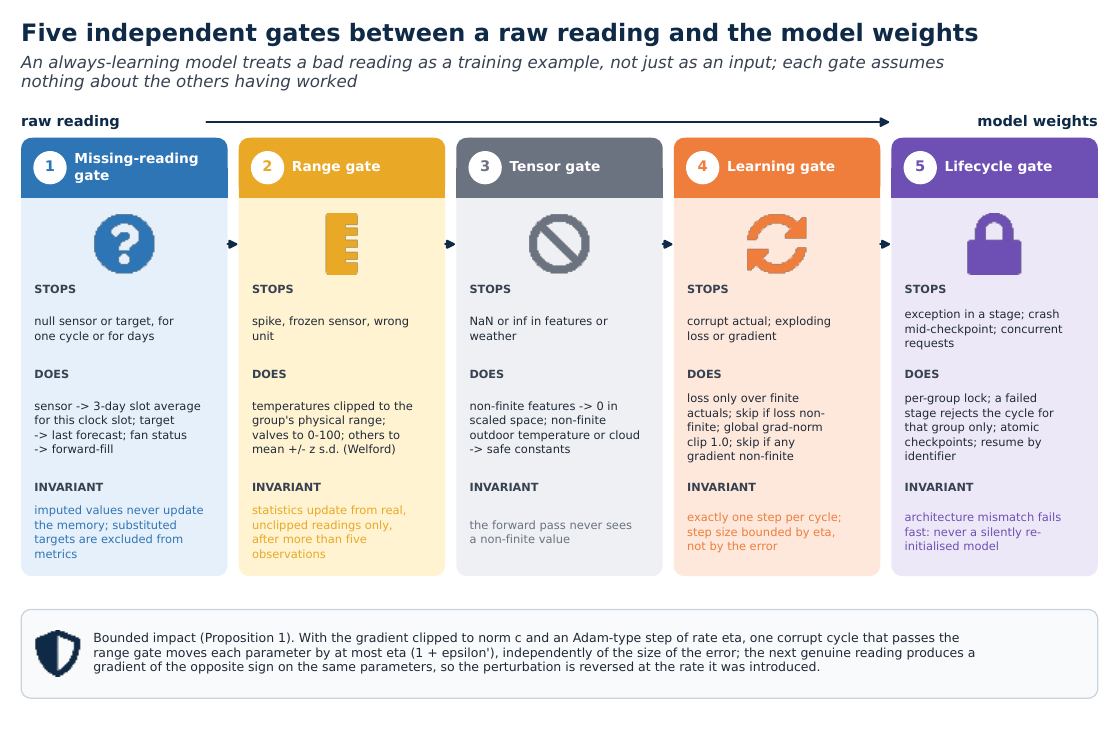}
\caption{The five gates a raw reading passes before it can influence the forecaster: for each, the fault it stops, what it does and the invariant it keeps; the bounded-impact property is stated beneath.}
\label{fig:gates}
\end{figure}

\textbf{Five gates and bounded impact.} An always-learning model treats a bad reading as a bad training example, so five independent gates stand between a reading and the weights (Figure~\ref{fig:gates}): missing-reading, range, tensor (non-finite to zero), learning (finite-loss and finite-gradient guards, global gradient-norm clip at 1.0) and lifecycle (per-group lock, per-cycle rejection, atomic checkpoints, fail-fast on architecture mismatch). With the clip at $c$ and an Adam-type step of rate $\eta$, a corrupt cycle that passes the range gate moves each parameter by at most $\eta(1+\epsilon')$ regardless of the error's magnitude, and the next genuine reading produces a gradient of opposite sign on the same parameters, and the perturbation decays with the optimiser's momentum over the following cycles rather than persisting; the memory and statistics are untouched by construction \ev{Proposed}.

\begin{figure}[t]
\centering\includegraphics[width=\columnwidth]{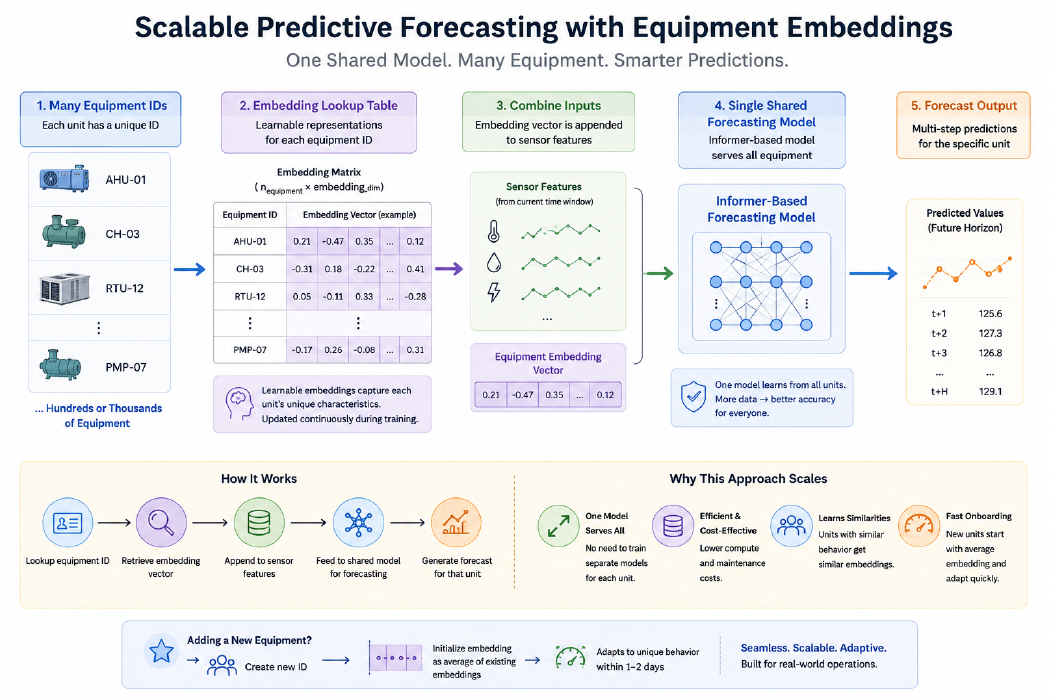}
\caption{Scalable forecasting with equipment embeddings: one shared model per group, one learned row per unit, new units initialised to the group mean (EEAI technical deck).}
\label{fig:emb}
\end{figure}

\textbf{Hand-off.} Every cycle the service returns, per unit, the next-cycle forecast, a six-step recursive horizon, a mean radiant temperature from a site-configurable linear model and the ISO 7730 PMV/PPD indices \cite{fanger1970} at the forecast state, plus the previous forecast's error as a trust signal. A rule-based setpoint-and-valve chain is retained as a verified fallback.

\section{Reinforcement-learning valve control}\label{sec:rl}
\textbf{MDP and reward.} Per unit, a continuing MDP at $\Delta t=5$\,min with a 21-feature observation normalised by fixed physical ranges (thermal sensors, humidity, energy, valve feedback, action history, a PID hint, the setpoint slot, time encodings, weather), action $a_t\in[0,100]\,\%$, $\gamma=0.98$, and \texttt{done} always false (gaps and time limits are truncations). The reward on the successor state is
\begin{equation}
R_t = w_cC_r(e_{t+1}) + w_eE_r(E_{t+1}) - c_s\tfrac{|a_t-a_{t-1}|}{\Delta_{\max}} - w_v\tfrac{\max(0,a_t)}{100},
\end{equation}
with a tiered comfort penalty $C_r\in[-1,0]$ (dead band below 0.5\,$\degC$; $-1$ at 2\,$\degC$) and a hinge energy term $E_r=-\alpha(E-T)$ for $E\ge T$ and $+\beta(T-E)$ otherwise with $\beta=0$, because rewarding under-consumption creates a degenerate optimum in which a closed valve earns reward while comfort collapses. The deployed configuration (policy ``7E'') uses $w_c/w_e=0.4/0.6$, $\alpha=1.5$ (2.5 for AHUs), target percentile 5 over active steps, FCUs comfort-only \ev{Proposed, Simulated}.

\begin{figure}[t]
\centering\includegraphics[width=\columnwidth]{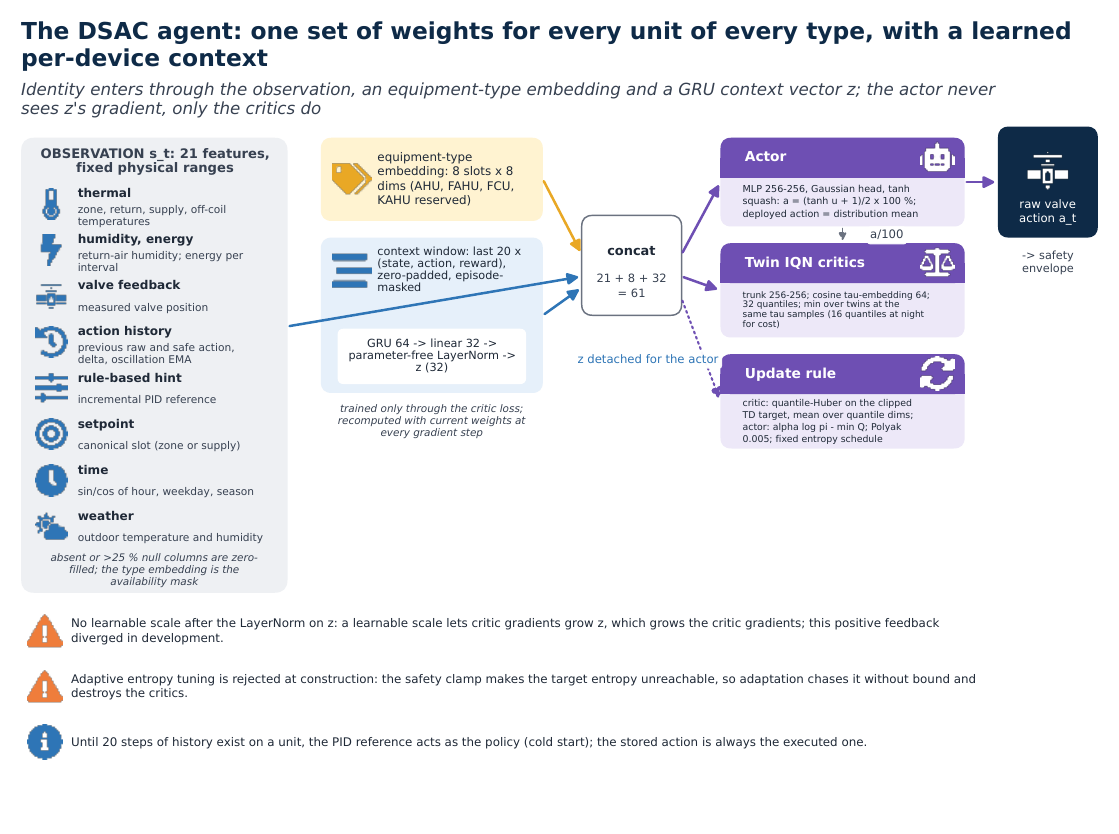}
\caption{The DSAC agent with learned per-device context: observation, type embedding, GRU context, actor, twin IQN critics and update rule, with the design notes recording failure modes found in development.}
\label{fig:agent}
\end{figure}

\textbf{Agent.} One set of weights serves every unit of every type (Figure~\ref{fig:agent}): a type embedding ($8\times8$), a GRU context encoder \cite{cho2014gru,rakelly2019pearl} over the last twenty (state, action, reward) steps with \emph{parameter-free} layer normalisation (a learnable scale after the normalisation lets critic gradients grow $z$, which grows the critic gradients, a positive feedback that diverged in development), a tanh-squashed Gaussian actor with the context detached, and twin IQN critics \cite{dabney2018iqn} evaluated at the same $\tau$ samples for the minimum. The critic target is clipped to $\pm q_{\text{clamp}}=|r|_{\max}/(1-\gamma)$; the quantile-Huber loss is reduced by the mean over both quantile dimensions; entropy follows a fixed cosine schedule because adaptive tuning chases a target entropy the safety clamp makes unreachable.

\textbf{Structural safety envelope.} Every command passes the rate clamp $\bar a_t=\clip(\bar a_{t-1}+\clip(a_t-\bar a_{t-1},\pm10),0,100)$, where $\bar a_{t-1}$ is the last executed command regardless of who issued it; on the device a dwell guard (hold three ticks, 2\,\% deadband) and a shadow/active state machine (promote after at least 5\,h and a predicted violation rate below 0.40, or by operator; demote automatically at a measured rate of 0.60) follow, and BMS writes require a \emph{live} registry status. None of these layers has a learnable parameter \cite{alshiekh2018shielding}. Every abnormal path writes to \texttt{errors.log}, so ``zero errors'' is auditable as an empty file; per-unit state is persisted atomically every tick, so a \texttt{kill -9} resumes without dropping or duplicating a transition.

\begin{figure*}[t]
\centering\includegraphics[width=0.95\textwidth]{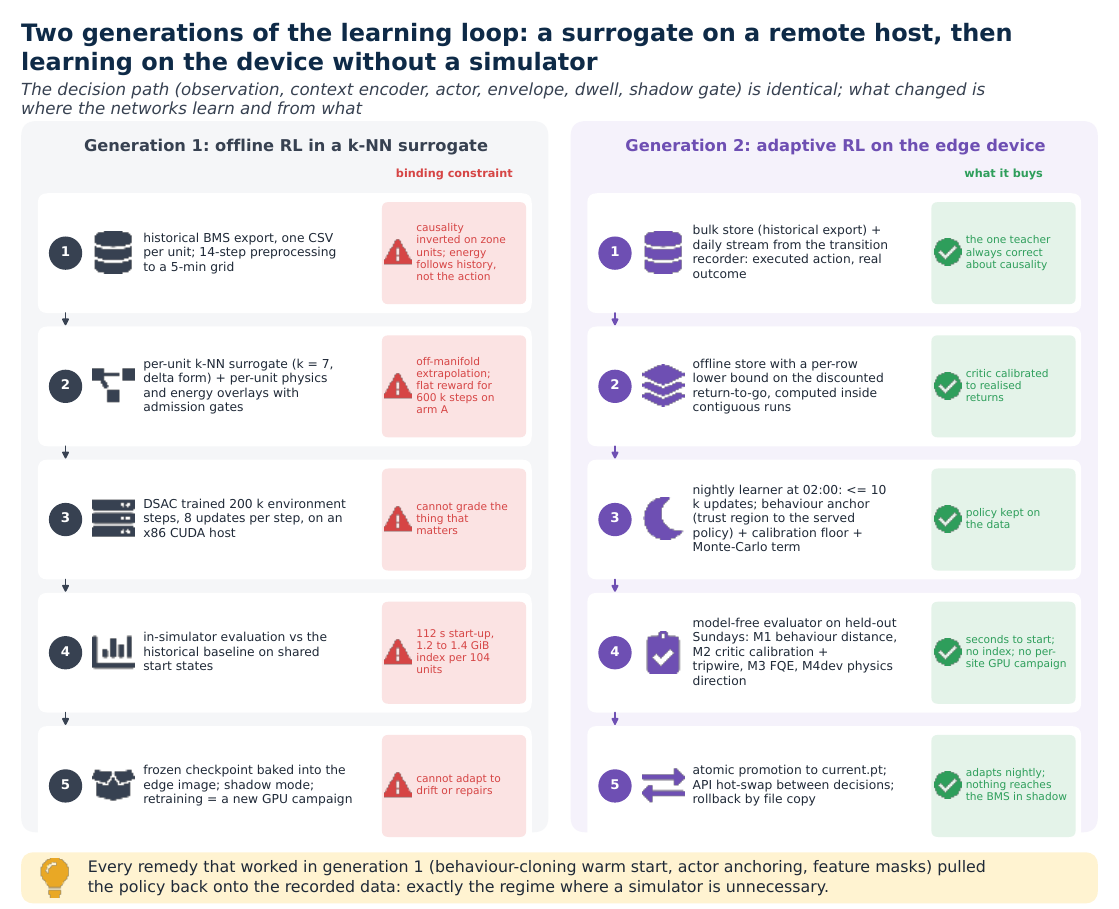}
\caption{The two generations of the learning loop, with the binding constraint of each generation-1 step and what each generation-2 step buys. The decision path is identical.}
\label{fig:v1v2}
\end{figure*}

\textbf{Generation 1 and why it was retired.} With no physics simulator and no licence to explore, v1 built a per-unit $k$-NN surrogate (delta form, $k=7$) from the BMS log and trained DSAC offline on a CUDA host (Figure~\ref{fig:v1v2}). On some zone-side units pure lookup reproduced the closed-loop correlation (hot room, open valve) rather than the causal effect, on most it had no valve authority at all, and its energy channel followed history rather than the action, so per-unit physics and energy overlays with their own admission gates were added. Five findings converged (Section~\ref{sec:res}): causal fidelity was fragile and unit-specific; a policy trained from scratch lived off the data manifold, and every remedy that worked pulled it back onto the recorded data, exactly where a simulator is unnecessary; the surrogate could not grade the thing that matters; it rebuilt a 1.2 to 1.4\,GiB index at every start (112\,s at 104 units) and needed a GPU campaign per site; and a frozen checkpoint could not adapt.

\textbf{Generation 2: adaptive RL on the device.} The surrogate is replaced by the site's recorded transitions, the one teacher always correct about causality. A transition recorder closes $(s_k,a_k,r_k,s_{k+1})$ with the executed command when the next reading arrives; an offline store keeps contiguous runs with a per-row \emph{lower bound} on the discounted return-to-go ($G_k=R_k+\gamma G_{k+1}$ with the unobserved tail valued at $-q_{\text{clamp}}$); a nightly learner at 02:00 site time, at reduced priority on four threads, performs up to 10\,k updates on a mixture of the day's rows and history; an evaluator scores the candidate on held-out Sundays; and promotion replaces \texttt{current.pt} atomically, which the API hot-swaps between decisions and names on its health endpoint, with rollback by file copy. Three terms answer the failure modes of plain DSAC on logs: a TD3+BC-style anchor $\beta_{\text{bc}}((\pi(s)-a_{\text{anchor}})/0.10)^2$ \cite{fujimoto2021td3bc} whose target is the deployed policy's own action (a trust region); a Cal-QL-style calibration floor and Monte-Carlo term \cite{nakamoto2023calql} regressing $\E[Q](s,a_{\text{rec}})$ to the return bound on rows with at least 194 steps left in their run; and removal of the hard clamp on \emph{predicted} quantiles, which has a dead gradient once saturated. Start-up drops from 112\,s to seconds.

\textbf{Coupling ladder.} The forecaster reaches the controller at three levels: operational (live; the forecasting block is logged and reconciled weekly), objective (implemented, off in 7E; the MRT model replaces the static setpoint in the reward) and observation (designed, not enabled; the forecast enters the policy's observation after a contract revision and a train/serve consistency proof). We state this because ``forecast-driven control'' is ambiguous without it.

\section{Counterfactual baseline}\label{sec:mv}
A building runs one control system at a time, so savings are counterfactual: $S_d=\hat E_d-E_d$ with $\hat E_d=g_\psi(x_d)$ fitted on BMS-operated days only (IPMVP Option C \cite{ipmvp2022}). Features are built by a recipe saved with the model; two are fitted inside every fold: a balance-point cooling-degree feature $\max(T_{\text{mean}}-T_b,0)$ with $T_b$ by grid search (20.25\,$\degC$ selected), which turns any linear model into an ASHRAE change-point model \cite{kissock2003imt}, and an enthalpy residual after OLS on temperature and dew point, whose fitted coefficients $(-4.686,1.197,1.917)$ match psychrometric theory $(-4.762,1.196,1.922)$. A 58-column pool is pruned (leakage guard, near-constant, $|r|>0.95$, permutation importance with a must-keep list, target-aware VIF $<10$) to 9 features for linear models. All preprocessing is one scikit-learn pipeline refitted per fold \cite{pedregosa2011sklearn,kaufman2012leakage}; validation is five expanding windows plus a 43-day validation split \cite{bergmeir2012cv}. Six families (OLS, Prophet with flat growth, random forest, histogram boosting, XGBoost, LightGBM; 295 candidates) are ranked by a robust objective $J=\text{mean}_w\text{CV(RMSE)}_w+\text{sd}_w\text{CV(RMSE)}_w+\text{mean}_w|\text{NMBE}_w|$ over six chronological windows, so a model wins only if accurate, consistent across seasons and unbiased. Acceptance gates on the validation split (accuracy $\ge0.85$, CV(RMSE) $\le15\,\%$, $|$NMBE$|\le5\,\%$, VIF $<10$) decide packaging; reporting gates on test periods never trigger retraining. Totals carry a 7-day moving-block bootstrap interval; exact Shapley values \cite{lundberg2017shap} explain every day. The package is served on the gateway as a stateless REST container with a data contract, self-checks, quality flags and an audit log; the runtime stores $S_d$ keyed by (date, model version).

\begin{figure*}[t]
\centering\includegraphics[width=0.9\textwidth]{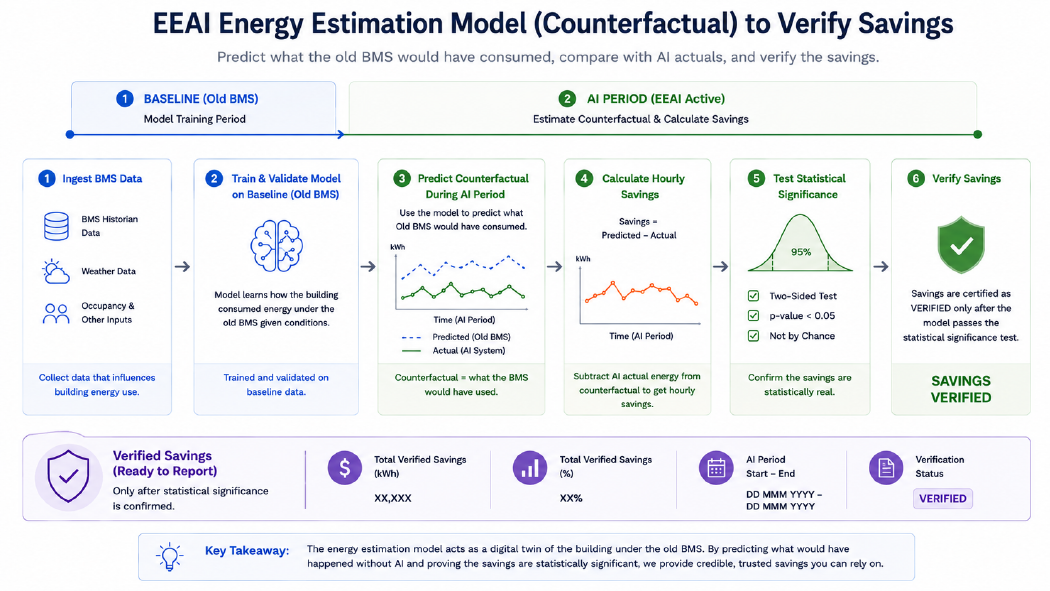}
\caption{From baseline to verified savings: the end-to-end M\&V workflow of the baseline component (EEAI technical deck).}
\label{fig:mvflow}
\end{figure*}

\section{Evaluation framework}\label{sec:eval}
\begin{figure}[t]
\centering\includegraphics[width=\columnwidth]{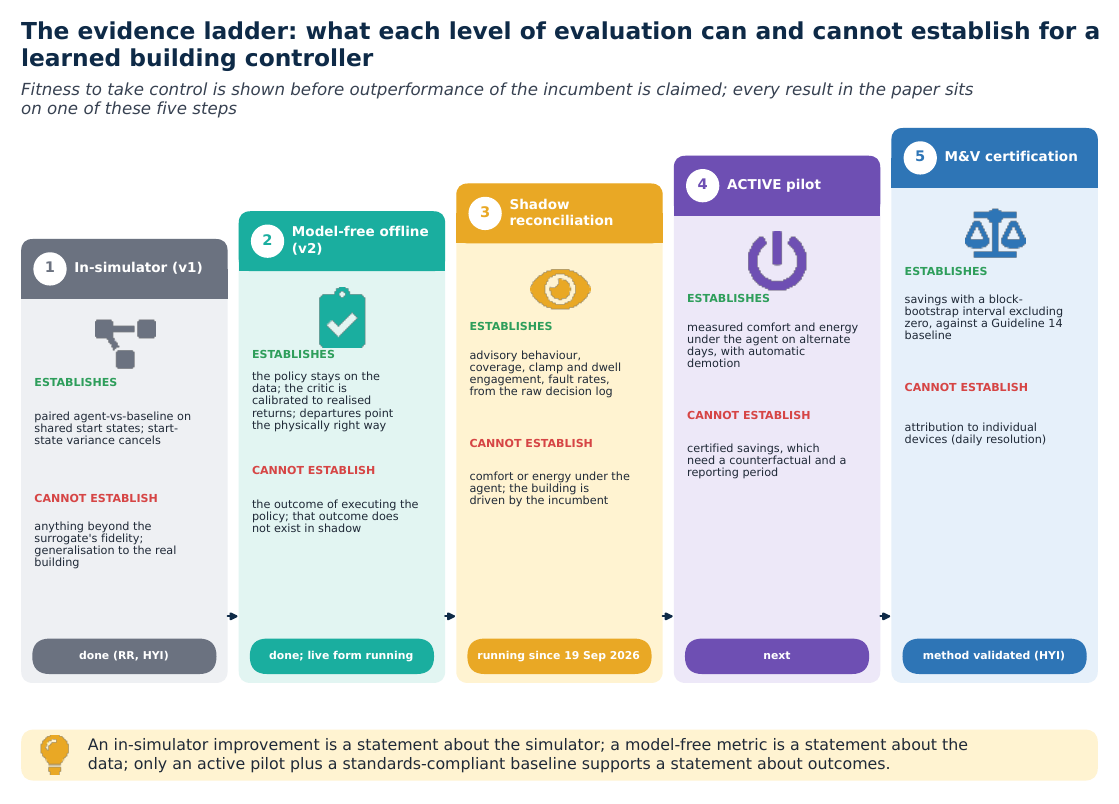}
\caption{The evidence ladder: fitness to take control is shown before outperformance is claimed.}
\label{fig:evidence}
\end{figure}

Five levels of evidence establish different things (Figure~\ref{fig:evidence}) \ev{Proposed}: in-simulator paired comparison on shared starts (a statement about the surrogate); model-free offline metrics on held-out days (a statement about a policy's relation to the data); shadow reconciliation (about the advisory); an active pilot (about outcomes); and M\&V certification (about savings). For v2 we define four model-free metrics: M1 behaviour distance (median and P90 of $|a_\pi-a_{\text{BMS}}|$, share within 10\,pp, saturation); M2 critic calibration ($|\E[Q](s,a_{\text{BMS}})-$realised return$|$ on long-horizon rows, with a tripwire at MAE $>5$ that voids the candidate's value metrics); M3 fitted Q evaluation \cite{le2019fqe}, directional only; and M4dev, the share of departures of at least 5\,pp from the incumbent that point the physically right way (the unconditioned form is confounded with M1). Savings are certified by a six-step protocol ending in a block-bootstrap interval that excludes zero, and Guideline 14's uncertainty relation $\Delta E/E = t\cdot1.26\cdot\mathrm{CV}\sqrt{(n/n')(1+2/n)/m}/F$ sizes the active pilot: at the measured 8.6\,\% CV(RMSE), with $n=365$ and $\rho=0.70$, a 10\,\% saving is resolved to 50\,\% relative uncertainty at 68\,\% confidence in about four weeks and a 5\,\% saving in about four months.

\section{Results}\label{sec:res}
\begin{figure*}[t]
\centering\includegraphics[width=0.98\textwidth]{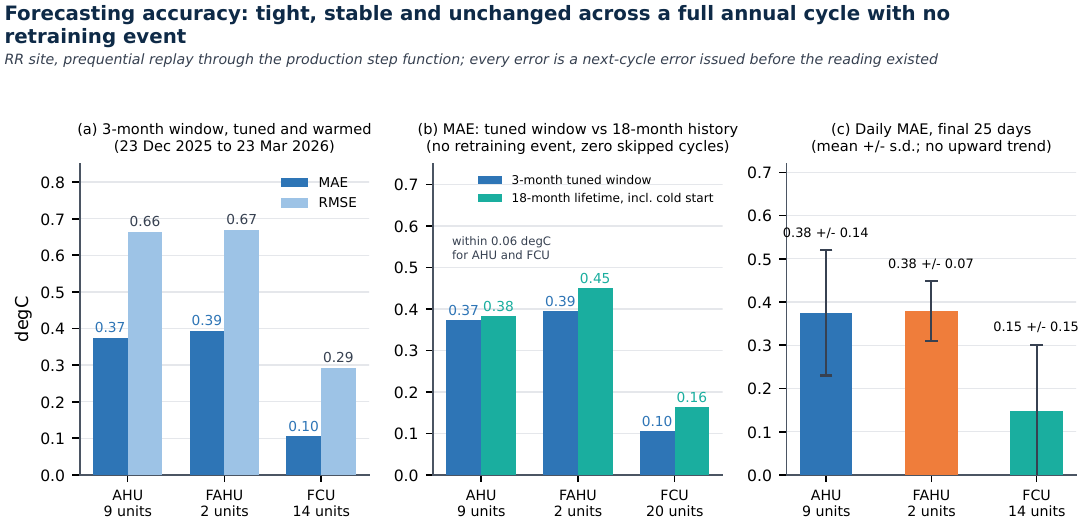}
\caption{Forecasting accuracy. (a) Three-month replay, MAE and RMSE. (b) Tuned window versus the full eighteen-month history with no retraining event. (c) Daily MAE, mean and s.d., final 25 days.}
\label{fig:fc}
\end{figure*}

\textbf{Forecasting} \ev{Measured}. RR exports from September 2024 to March 2026 were replayed through the service as live requests. A staged search (22 replays per group over a three-month window) found that the smallest architecture wins or ties everywhere, dropout hurts in every group and learning rate is the dominant knob, consistent with single-instance updates favouring low-capacity models. From random weights, most of the gain arrives inside the first week (cumulative MAE after about two days, two weeks and 3.5 months: AHU 2.44, 1.04, 0.71; FAHU 1.28, 0.84, 0.66; FCU 0.61, 0.26, 0.21\,$\degC$), the 201-unit FCU group fastest because every unit's reading trains the shared model. Tuned and warmed, the three-month replay gives the figures of Table~\ref{tab:fc} and Figure~\ref{fig:fc}, with all nine AHUs between 0.33 and 0.50\,$\degC$ and all fourteen FCUs between 0.09 and 0.15; the continuous eighteen-month replay, including cold start and every seasonal transition, gives lifetime MAE of 0.383, 0.450 and 0.163\,$\degC$ with zero skipped cycles and no retraining event, and the daily MAE over the final 25 days shows no upward trend. Three recorded one-cycle spikes (5.1, 3.7, 3.3\,$\degC$) produced next-cycle errors of 0.48, 6.5 (a genuinely noisy period) and 0.45\,$\degC$ and settled within 1, 5 and 9 cycles with no lasting offset.

\begin{table}[t]
\caption{Forecasting accuracy, three-month prequential replay (RR, 23 Dec 2025 to 23 Mar 2026).}
\label{tab:fc}
\vskip 0.1in
\centering\footnotesize
\begin{tabular}{@{}lrrrrr@{}}
\toprule
Group & Units & Cycles & MAE ($\degC$) & RMSE ($\degC$) & sMAPE \\
\midrule
AHU & 9 & 270\,568 & 0.373 & 0.664 & 1.63\,\% \\
FAHU & 2 & 82\,821 & 0.394 & 0.669 & 1.83\,\% \\
FCU & 14 & 558\,218 & 0.105 & 0.293 & 0.45\,\% \\
\bottomrule
\end{tabular}
\vskip -0.1in
\end{table}

\textbf{RL, generation 1} \ev{Simulated}. A lineage of preregistered runs on RR moved the objective from the specification's defaults to 7E; the single-unit paired result reported internally (comfort violations 57.4\,\% to 42.3\,\%, energy 90.2\,\% of the historical rollout) is an in-simulator quantity. The HYI arms exposed the surrogate: arm A's reward stayed flat for 600\,k steps because the surrogate was not answering the action; arm A$'$ spent 11 to 51\,\% of steps beyond 2\,$\degC$ with no gradient; arm A$''$'s critic extrapolated to $-34\,000$ until the actor objective was clamped and its behaviour-cloning term was rescaled by two orders of magnitude; arm A4 met its preregistered gates. The fidelity harness found valve and energy moving in opposite directions on 26 of 45 units ($r=-0.19$) under pure lookup against 0.65 to 0.97 in the raw data, which a sign-constrained energy overlay corrected while cutting six-hour energy error about six-fold.

\textbf{RL, shadow trial and generation 2} \ev{Measured}. Since 19 September 2026 the frozen 7E image serves 104 RR units in SHADOW under a 2\,GB, one-CPU limit: 172 to 188\,ms decision round trip for 95 live units, 6.3\,ms policy forward, 1.43\,GiB steady RSS, 112\,s start-up (k-NN index). First live observations: the advisory tracks zone temperature the right way ($r=+0.55$), output is bang-bang on the zero-shot FCU class, the clamp barely engages (5.6\,pp mean), 23\,\% of FCU cycles are fan-off where the export had FCUs always on, and energy is posted null pending a meter repair, so the energy verdict is not yet computable. Two independent read-only gates were verified; zero write-back rows trace to an RL decision. In the v2 replay experiment on the MIC-713 (sixty virtual days in 2.5 days), Arm 0 (the specification as written) reproduced the failure modes the new terms address, with its critic pinned at the clamp and its policy drifting 49 to 90\,pp from the incumbent, and plain TD propagated value scale too slowly ($Q=-2$ against realised returns of $-30$ after 55\,k steps), which fixed the Monte-Carlo weight at 3. The live v2 form starts from 7E with the anchor on 7E's own action, trains at most 10\,k steps per night (266\,ms per step with 16 quantiles) and has preregistered four-week criteria (median distance to 7E $\le5$\,pp, saturation $<10\,\%$, critic error falling, M4dev not below week one, zero errors); interim values are not reported here.

\begin{figure}[t]
\centering\includegraphics[width=\columnwidth]{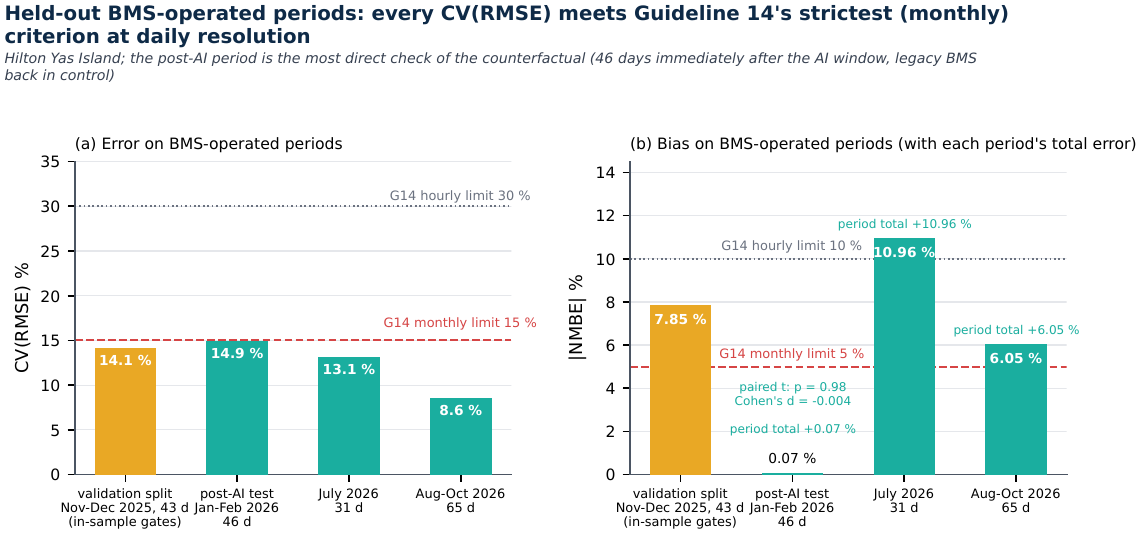}
\caption{Baseline error and bias on held-out BMS-operated periods against Guideline 14.}
\label{fig:mv}
\end{figure}

\textbf{Baseline} \ev{Measured}. On HYI the change-point regression and Prophet tie (robust score 17.69 versus 17.71) and beat the best tree ensemble by about 23\,\% on every component of the score; the linear model is deployed for interpretability and zero runtime dependencies. On the 43-day validation split it passes accuracy (0.859), CV(RMSE) (14.12\,\%), VIF (2.71) and condition number (3.94) and fails bias (NMBE $+7.85\,\%$) because of a documented regime shift (26\,\% more energy per unit of peak enthalpy in December 2025 than a year earlier). On held-out BMS-operated periods (Figure~\ref{fig:mv}) it reproduces the 46-day post-AI period to $+0.07\,\%$ in total (paired $t$, $p=0.98$) and reaches CV(RMSE) of 8.6 to 13.1\,\% seven to eleven months after training; every CV(RMSE) satisfies Guideline 14's monthly criterion at daily resolution. Retraining with the growing non-AI history (607 days) cuts total error on a common held-out window from 6.05\,\% to 1.38\,\%, CV(RMSE) from 8.56 to 6.49\,\% and lifts $R^2$ from about zero to 0.40. Shapley attribution is dominated by cooling degree-days (52.5\,\%) and dew point (29.1\,\%), and every main effect carries the sign physics predicts.

\begin{figure}[t]
\centering\includegraphics[width=\columnwidth]{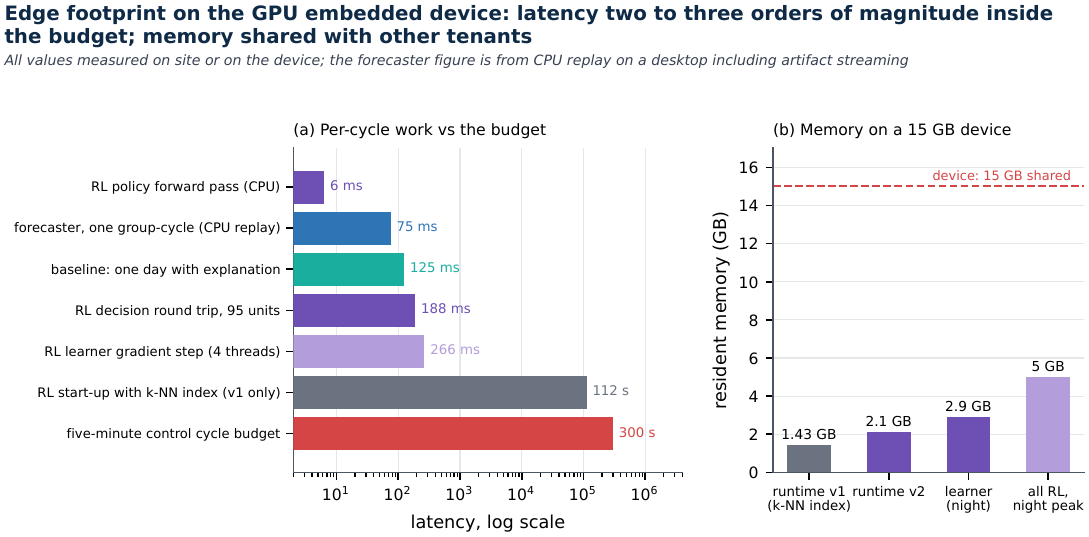}
\caption{Measured latency (log scale, against the five-minute budget) and memory (against the 15\,GB device) on the MIC-713 gateway.}
\label{fig:footprint}
\end{figure}

\textbf{Edge footprint} \ev{Measured}. Figure~\ref{fig:footprint}: forecaster 50 to 75\,ms per group-cycle (CPU replay), RL 6.3\,ms forward and 188\,ms round trip, baseline 125\,ms per day with explanation, learner 266\,ms per gradient step (10\,k steps in about 45 minutes; 100\,k bulk in 7.4\,h); memory 2.1\,GB runtime, 2.9\,GB learner, about 5\,GB RL peak at night on a 15\,GB device shared with the forecaster.

\section{Systems lessons}\label{sec:lessons}
\textbf{Train/serve consistency must be testable.} Context windows are rebuilt at sample time with an episode-boundary mask to reproduce exactly the inference-time history; the recorder and the store share semantics; tests assert bit-exact preprocessing and byte-identical same-seed training. A one-step-early reward timing and a label/contract drift (78 of 104 units reporting ACTIVE while the API kept the command null) were caught by tests, not in production. \textbf{Data semantics are load-bearing.} Per-column grid merging (whole-row deduplication halved usable data with mixed 1- and 5-minute sampling), median rather than zero filling of sparse gaps, action-history features that do not fabricate valve moves across gaps, and a setpoint slot wide enough for zone and supply targets all changed results materially. \textbf{Fail loud; never silently re-initialise.} Missing configuration keys raise by name; an architecture mismatch at load refuses to start, because a re-initialised always-on learner serves noise while appearing healthy. \textbf{The image tag is the model version.} Checkpoints are baked into tags that are never reused; the in-image suite (178 tests) proves the training and runtime images' divergent numerical pins agree. \textbf{Preregister the question and the gate.} Every arm is a configuration directory with an orchestrator asserting it before training and running the evaluation chain in a fixed order; this discipline is what let surrogate defects be recognised rather than tuned around. \textbf{Evaluate the deployed object.} Replay through the production step function, a runtime simulator against the live baseline API, and nightly scoring of the exact file that will be hot-swapped are instances of one rule.

\section{Limitations and roadmap}\label{sec:limits}
Both controller generations have operated only in shadow; the first reporting window's saving awaits a live meter and the active pilot. The forecast is logged but not yet in the policy's observation. The forecaster emits point forecasts, is pulled once by a corrupt reading inside the range gate and can ring on fast targets. The 7E policy never trained on zone-only FCUs, and the export's missing fan status is a known flaw. The baseline covers one site and under two annual cycles at daily resolution. The roadmap follows the evidence ladder: an active pilot with alternate-day comparison and automatic demotion, sized by the Guideline 14 relation; observation coupling with a consistency proof; prediction intervals and adaptive per-unit learning rates; adaptive-window drift monitoring \cite{bifet2007adwin}; an hourly baseline with device attribution; multi-site transfer; and a supervisory language-model agent over the audit trail and configuration tools that triages, reports and assists commissioning but never issues a valve command, in line with the deployment-readiness evidence \cite{neubauer2026llmhvac}.

\section{Conclusion}
Three small, embodied, self-evaluating and governed agents on one gateway replace the collect-train-deploy-retrain pattern: a forecaster that learns every cycle and holds its accuracy across a year with no retraining event, a controller that learns nightly from the building's own transitions behind an envelope no gradient can soften, and a baseline that turns actions into certifiable savings. We have stated exactly what the evidence establishes, fitness to take control, and what remains, the active pilot the stack already carries the mechanisms to run.

\bibliography{main}
\bibliographystyle{mlsys2025}

\end{document}

%% file: glossary.tex
\newacronym{Al-Jazeera}{Al-Jazeera}{Al-Jazeera Satellite Channel}
\newacronym{AraGPT-2}{AraGPT-2}{Arabic Generative Pre-trained Transformer 2}
\newacronym{AraGPT-3}{AraGPT-3}{Arabic Generative Pre-trained Transformer 3}
\newacronym{ArSAS}{ArSAS}{Arabic Speech-Act and Sentiment}
\newacronym{PATB}{PATB}{Penn Arabic Treebank}
\newacronym{BBC}{BBC}{British Broadcasting Corporation}
\newacronym{BERT}{BERT}{Bidirectional Encoder Representations from Transformers}
\newacronym{BPE}{BPE}{byte-pair encoding}
\newacronym{mBERT}{mBERT}{multi-lingual BERT}
\newacronym{RoBERTa}{RoBERTa}{A Robustly Optimized BERT Pre-training Approach}
\newacronym{CA}{CA}{Classical Arabic}
\newacronym{CC}{CC}{Common Crawl}
\newacronym{CC-100}{CC-100}{Common Crawl 100}
\newacronym{CNN}{CNN}{Cable News Network}
\newacronym{DA}{DA}{Dialectal Arabic}
\newacronym{EBIT}{EBIT}{Earnings Before Interest and Taxes}
\newacronym{EAPCOUNT}{EAPCOUNT}{English-Arabic Parallel Corpus of the United Nations Texts}
\newacronym{GenAI}{GenAI}{Generative AI}
\newacronym{eVal}{eVal}{EnterpriseVal}
\newacronym{GPT-1}{GPT-1}{Generative Pre-trained Transformer 1}
\newacronym{GPT-2}{GPT-2}{Generative Pre-trained Transformer 2}
\newacronym{GPT-3}{GPT-3}{Generative Pre-trained Transformer 3}
\newacronym{GPT}{GPT}{Generative Pre-trained Transformer}
\newacronym{GPU}{GPU}{Graphics Processing Unit}
\newacronym{HPL}{HPL}{High-Performance Linpack} 
\newacronym{IR}{IR}{Information Retrieval} 
\newacronym{KSUCCA}{KSUCCA}{King Saud University Corpus of Classical Arabic}
\newacronym{LM}{LM}{Language model}
\newacronym{LLM}{LLM}{Large Language Model}
\newacronym{LSTM}{LSTM}{Long-Short Term Memory}
\newacronym{ME}{ME}{Middle East}
\newacronym{MGB-2}{MGB-2}{Multi-Genre Broadcast-2}
\newacronym{MSA}{MSA}{Modern Standard Arabic}
\newacronym{MEGLN}{MEGLN}{MSA, Egyptian, Gulf, Levant \& North African}
\newacronym{MRC}{MRC}{Machine Reading Comprehension}
\newacronym{MT}{MT}{Machine Translation}
\newacronym{NN}{NN}{Neural Network}
\newacronym{NQ}{NQ}{Natural Questions}
\newacronym{NLP}{NLP}{Natural Language Processing}
\newacronym{OpenITI}{OpenITI}{Open Islamicate Texts Initiative}
\newacronym{OSAC}{OSAC}{Open-Source Arabic Corpora}
\newacronym{OSCAR}{OSCAR}{Open Super-large Crawled ALMAnaCH coRpus}
\newacronym{OSIAN}{OSIAN}{Open Source International Arabic News}
\newacronym{POS}{POS}{Part-of-speech}
\newacronym{PTB}{PTB}{Penn Tree Bank}
\newacronym{QA}{QA}{Question Answering} 
\newacronym{RNN}{RNN}{Recurrent Neural Network}
\newacronym{RT}{RT}{Russia Today}
\newacronym{SM}{SM}{Sentence Match}
\newacronym{SQuAD}{SQuAD}{Stanford Question Answering Dataset}
\newacronym{TPU}{TPU}{Tensor Processing Unit}
\newacronym{TFRC}{TFRC}{TensorFlow Research Cloud}
\newacronym{KSU}{KSU}{King Saud University}
\newacronym{WER}{WER}{Word Error Rate}
\newacronym{XNLI}{XNLI}{Cross-lingual Natural Language Inference}
\newacronym{HARD}{HARD}{Hotel Arabic Reviews Dataset}